# InceptB: A CNN Based Classification Approach for Recognizing Traditional Bengali Games

Mohammad Shakirul Islam[a*], Ferdouse Ahmed Foysal[a], Nafis Neehal[a], Enamul Karim[a]
Syed Akhter Hossain[a]

[a]Department of Computer Science and Engineering, Daffodil International University, Dhaka-1207, Bangladesh

**Abstract**

Sports activities are an integral part of our day to day life. Introducing autonomous decision making and predictive models to recognize and analyze different sports events and activities has become an emerging trend in computer vision arena. Albeit the advances and vivid applications of artificial intelligence and computer vision in recognizing different popular western games, there remains a very minimal amount of efforts in the application of computer vision in recognizing traditional Bangladeshi games. We, in this paper, have described a novel Deep Learning based approach for recognizing traditional Bengali games. We have retrained the final layer of the renowned Inception V3 architecture developed by Google for our classification approach. Our approach shows promising results with an average accuracy of 80% approximately in correctly recognizing among 5 traditional Bangladeshi sports events.

## 1. Introduction

Sports are the wellsprings of amusement. They are the sources of excitements and pleasantries in our dull and monotonous life. Application of Computer Vision (CV) and Artificial Intelligence (AI) is becoming popular day by day in analyzing different sports activities.

By introducing automation and predictive modelling, it increases the efficiency in different decision making procedures involved with various sports. Along with decision making, application of computer vision in recognizing different sports activities and events has lately become very popular also. Although, a lot of progress has been made in the arena of applying AI and Computer Vision in different western sport activities, the contribution in Traditional Bengali games remains somewhat insignificant in amount.

There are some traditional games in Bangladesh. Most people of present generation doesn't know about this games. They don't know the name by which the game is called. Kabaddi is the national sport of Bangladesh. Except this, there are some popular sports like cricket, football, hockey, volleyball, danguli, kabadi, kanamachi, latthi khela, nouka baich, boxing, cycling, long jump, handball, golf, badminton, running, swimming and so on. Every sport is different from each other. Among those, initially we want to recognize the Traditional Bangladeshi games, which are peculiar to our country, namely – "Danguli", "Kabadi", "Kanamachi", "Latthi Khela" and "Nouka Baich" from random images. We have developed a dataset containing images of each activity of the sports. In the Cenozoic Era we are attracted with modern foreign games like cricket, football etc. But we should know about our sports culture. So, our work serves from social perspective too.

Our main goal is to classify and recognize these game events accurately. We have deployed an inception v3 model which is developed by google to classify these sports event. We have used transfer learning method as we retrain only the final layer of Inception V3 model while hyper-parameters from other layers remain intact.

The remaining part of the paper is arranged in the following manner – Section 2 contains our background study, Section 3 contains our proposed methodology, Section 4 contains our model's performance evaluation, Section 5 contains the Result Discussion, Section 6 contains the future work and Section 7 contains the concluding remarks.

## 2. Literature Review

Till now, it has been found that there are a few works done on sports activity recognition. A few works done on various sports using deep learning[5], ImageNet, CNN. A research group from china used inception-v3 [3] on Oxford-17 and Oxford-102 flower dataset.Compared with any object classification methods, convolutional neural network [7],[8] use multilayer convolution to extract features and combine the features automatically. It has a higher recognition rate and a wider range of applications.

Inception-v3 [2] is one of the pre trained models on the TensorFlow [1]. It is a rethinking for the initial structure of computer vision after Inception-vI [5], Inception-v2 [4] in 2015. The Inception-v3 [2] model is trained on the ImageNet datasets, containing the information that can identify 1000 classes in ImageNet, the error rate of top-5 is 3.5%, the error rate of top-I dropped to 17.3%. Tensorflow [1] also provides detailed tutorials for us to retrain Inception's final Layer for new categories using transfer learning. Transfer learning is a new machine learning method which can use the existing knowledge learned from one environment and solve the other new problem which is different but has some relation with the old problem. For example, we can apply the knowledge learned from the car problem to the study of bicycle problem. Compared with the traditional neural network [9], it only needs to use a small amount of data to train the model, and achieve high accuracy with a short training time.

## 3. Proposed Methodology

This section focuses on the construction process of game classification model. The construction process of the model divided into few steps.

*3.1. Convolutional neural network*

Convolutional neural network[11],[6] (CNN) is network architecture for deep learning[5]. CNN are deep artificial neural networks [10] that are used primarily to classify images cluster them by similarity and perform object recognition[12] within scenes. A CNN is comprised of one or more convolutional layers and then followed by one or more fully connected layers as in a standard multilayer neural network .It learns directly from images. A CNN can be trained to do image analysis tasks including classification, object detection, segmentation and image processing.

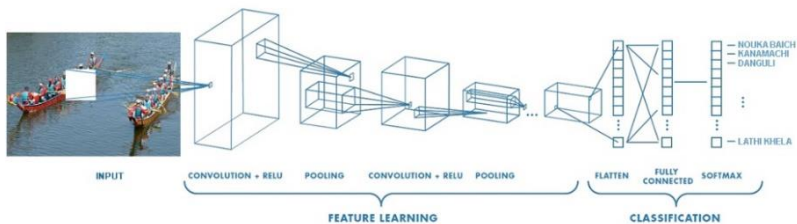

Fig.1. CNN Architecture.

CNN is made of several types of layer, like Convolutional Layer, Non-Linearity Layer, Rectification Layer, Rectified Linear Units (ReLU), Pooling Layer, Fully Connected Layer, Dropout Layer (Fig.1.).

- Convolutional Layer: The main task of the convolutional layer is to detect local conjunctions of features from the previous layer and mapping their appearance to a feature map.
- Non-Linearity Layer: A non-linearity layer in a convolutional neural network consists of an activation function that takes the feature map generated by the convolutional layer and creates the activation map as its output.
- Rectification Layer: A rectification layer in a convolutional neural network performs element-wise absolute value operation on the input volume.

- Rectified Linear Units (ReLU): The rectified linear units (ReLUs) are a special implementation that combines non-linearity and rectification layers in convolutional neural networks Pooling Layer. The pooling or down sampling layer is responsible for reducing the special size of the activation maps.
- Fully Connected Layer: The fully connected layers in a convolutional network are practically a multilayer perceptron (generally a two or three layer MLP) that aims to map the $m1(l-1) \times m2(l-1) \times m3(l-1)$ activation volume from the combination of previous different layers into a class probability distribution.
- Dropout Layer: Dropout is a technique used to improve over-fit on neural networks, you should use Dropout along with other techniques like L2 Regularization.
- Softmax: Softmax function calculates the probabilities distribution of the event over 'n' different events. In general way of saying, this function will calculate the probabilities of each target class over all possible target classes. Later the calculated probabilities will be helpful for determining the target class for the given inputs. In building neural networks softmax functions used in different layer level.

*3.2. Inception v3 model*

Inception v3 is the 2015 iteration of Google's Inception architecture (Fig.2.) for image recognition. Inception is a really great architecture and it's the result of multiple cycles of trial and error. The Inception[3] code uses TF-Slim, which seems to be a kind of abstraction library over TensorFlow[1] that makes writing convolutional nets easier and more compact. This model is trained on a subset of the ImageNet database, which is used in the ImageNet Large-Scale Visual Recognition Challenge (ILSVRC). The model is trained on more than a million images and can classify images into 1000 object categories, such as keyboard, mouse, pencil, and many animals. As a result, the model has learned rich feature representations for a wide range of images.

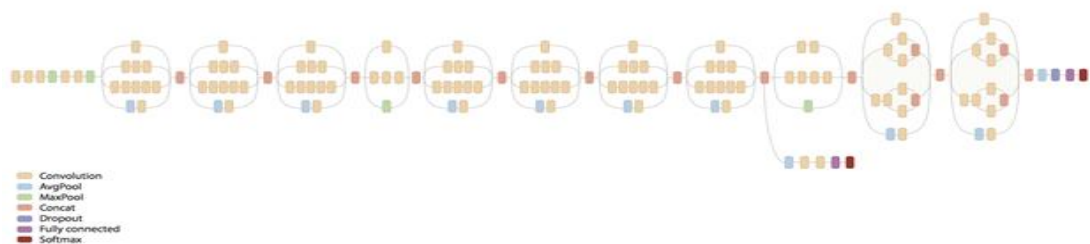

Fig.2. Architecture of Inception v3 model

There are three ways to use CNN [10]. We have used Transfer learning. Transfer learning is based on the idea, that the knowledge of solving one type of problem can be used to solve a similar problem .Using inception v3 model to classify images by retraining it is one kind of transfer learning.

*3.3. Dataset Collection*

This method used obtainable public data of some traditional Bangladeshi games. This dataset have 5 five classes of games images such as kabaddi, kanamachi, nouka baich, danguli etc. This dataset have total 3600 images 3000 images for training and 600 for testing. Each class have 600 images, for testing each test class have 120 images. To take a look in this dataset (Fig. 3.).

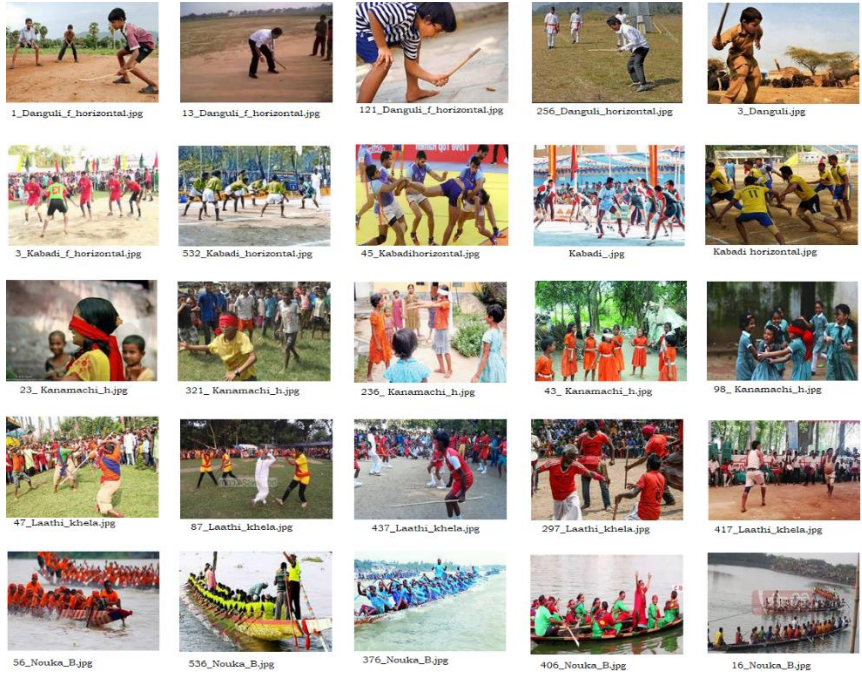

Fig. 3. The example dataset of Traditional Bangladeshi games.

### 3.4. Data Processing

#### 3.4.1 Data Augmentation

To avoid overfitting, we artificially expand the dataset. This data will create a few variance that can occur when someone else take new data from web or real life. After collecting data for each class we augment the dataset in 5 different methods, these methods given below:

- Rotate left -30 degree
- Rotate right +30 degree
- Flip horizontally about Y axis
- Shear by a certain amount
- Rotate left +90 degree

#### 3.4.2 Data Preparation (Resize)

After augmentation all the data have different heights and widths. Our model requires a fixed pixel for all data images. We resize all data in a fix resolution: the Leeds images to 200x150 pixels. We use RGB to get a good accuracy and cross entropy. We augment the dataset, this dataset is created by as we named this dataset InceptB.

This dataset contains 120 core images for each class, after augmentation and resize this dataset increase to 720 for each class, then we move 120 data from each class for test and rest of 3000 images was used to train our model. It is difficult to display all data in so we display five images for each class above in fig. 1.

*3.5. Proposed Inception Model*

A simplified picture of Inception-v3 model [2] is shown in figure 4. Inception-v3 [3] network model is a deep neural network, it is very difficult for us to train it directly with a low configured computer, it takes at least a few days to train it. Tensorflow [1] provides a tutorials for us to retrain Inception's final Layer for new categories using transfer learning. We use the transfer learning method which keep the parameters of the previous layer and remove the last layer of the Inception-v3 [2] model, then retrain a last layer. The number of output nodes in the last layer is equal to the number of categories in the dataset.

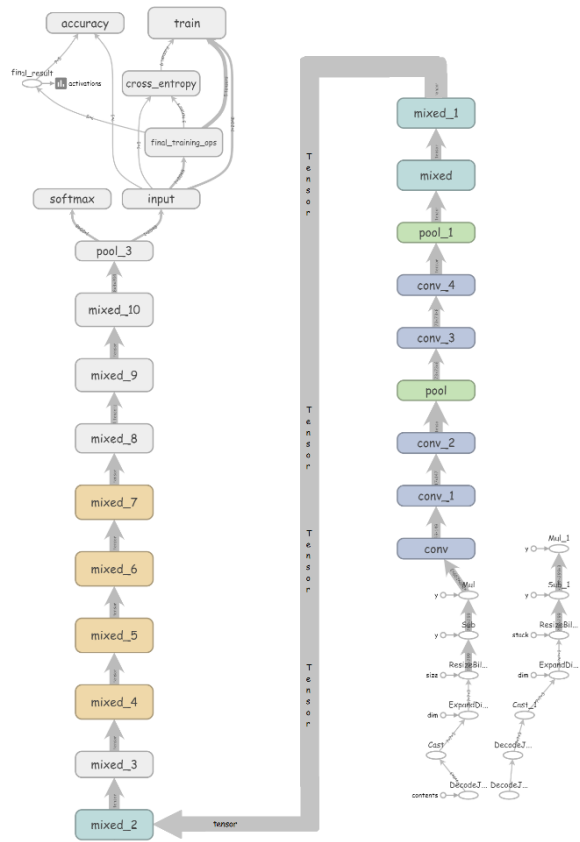

Fig. 4. A simplified picture of Inception-v3 from TensorBoard

*3.6. Training the model.*

After generating bottleneck files for all data, the main training of the final layer of the network started. The training operates efficiently by feeding the cache value for each image into the layer. The truth label for each image is also fed into the node labeled GroundTruth. After training we can see a series of step outputs, each one showing training accuracy, validation accuracy and the cross entropy. The training's objective is to make the cross entropy as small as possible, so we can tell if the learning is working by keeping an eye on whether the loss keeps trending downwards, ignoring the short-term noise.

By default this model script runs 4,000 training steps. Each step chooses 10 images at random from the training set, finds their bottlenecks from the cache, and feeds them into the final layer to get predictions. Those predictions are then compared against the actual labels to update the final layer's weights through a backpropagation process. As the process continues, we can see the reported accuracy improved. After all the training steps are completed, the script runs a final test accuracy evaluation on a set of images that are kept for testing purpose.

## 4. Performance Evaluation

The training accuracy (fig.5.a) shows the percentage of the data used in the current dataset that were labelled with the correct class and the validation accuracy (fig.5.a) shows the precision on a randomly-selected images from different class. The core difference is the training accuracy is based on images that the network has been able to learn from so the network can overfit to the noise in the data.

Cross entropy (fig.5.b) is a loss function which gives us a glimpse into how well the learning process is progressing. The training's objective is to make the loss as small as possible, so that we can tell if the learning is working by keeping an eye on whether the loss keeps trending downwards, ignoring the short-term noise.

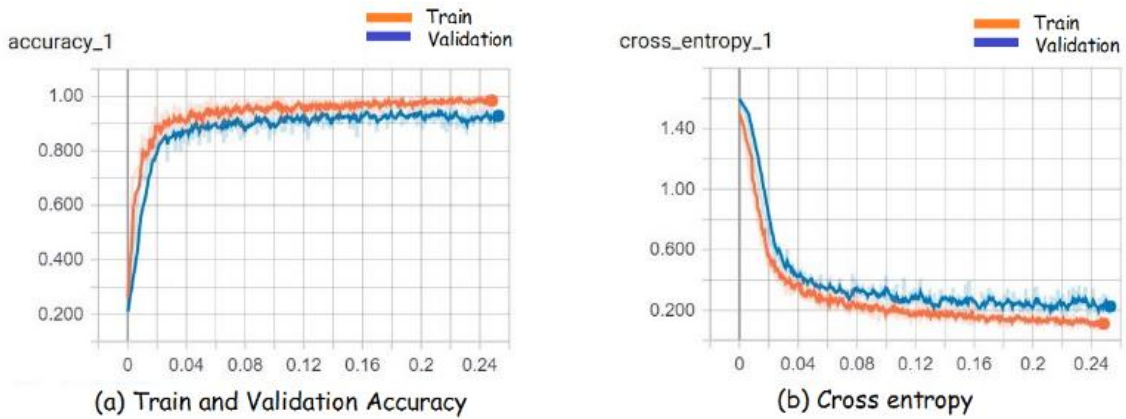

Fig.5. (a) Training and validation accuracy, (b) Cross Entropy

## 5. Result Discussion

This precision, Recall and F-Measure is calculated from working on a test dataset containing 600 images (120 images per class). From the Precision-Recall measurement, we can infer that our model has moderately high precision and high recall in an average (approximately 77.496% and 78.632% respectively) which indicates the decent performance of our classifier.

Table 1. Accuracy, precision, recall and f-measure of our Bangladeshi games dataset.

| Class | Precision | Recall | F-mesure |
|---|---|---|---|
| Danguli | 80.83% | 93.26% | 86.60% |
| Nouka Baich | 84.16% | 86.32% | 85.22% |
| Kabadi | 72.5% | 62.14% | 66.92% |
| Latthi Khela | 74.16% | 80.90% | 77.38% |
| Kanamachi | 75.83% | 70.54% | 73.08% |

From the Confusion matrix given below, we can infer that there is a very high number of True Positives and decently low number of False Positives that our model has outputted.

Table 2. Confusion Matrix

|  | Nouka Baich | Danguli | Kabadi | Kanamachi | Latthi khela |
|---|---|---|---|---|---|
| Nouka Baich | 101 | 0 | 10 | 2 | 7 |
| Danguli | 2 | 97 | 5 | 12 | 4 |
| Kabadi | 4 | 1 | 87 | 22 | 6 |
| Kanamachi | 4 | 5 | 16 | 91 | 4 |
| Latthi khela | 6 | 1 | 22 | 2 | 89 |

## 6. Future Work

We have developed this classification approach using Google's Inception V3 model which is giving us decent accuracy. But in future, we want to develop a CNN model from scratch in order to improve the accuracy.

## 7. Conclusion

This paper has used Google's renowned Inception-v3 model of Tensorflow platform along with the transfer learning technology to train a CNN Model on Traditional Bangladeshi Game image dataset for recognition purpose. The result we've achieved is really promising. Hopefully this approach will be pursued and developed in future as a part of further contributions in our Bangladeshi Society.